\documentclass[10pt,twocolumn,letterpaper]{article}

\usepackage{cvpr}              

\usepackage[dvipsnames]{xcolor}

\usepackage{multirow}
\usepackage{utfsym}
\usepackage{arydshln}
\usepackage{float}
\definecolor{cvprblue}{rgb}{0.21,0.49,0.74}
\usepackage[pagebackref,breaklinks,colorlinks,citecolor=cvprblue]{hyperref}

\def\paperID{60} 
\def\confName{CVPR}
\def\confYear{2024}

\title{HMANet: Hybrid Multi-Axis Aggregation Network for Image Super-Resolution}

\author{Shu-Chuan Chu$^1$, Zhi-Chao Dou$^1$, Jeng-Shyang Pan$^{2,*}$, Shaowei Weng$^3$, Junbao Li$^4$\\
$^1$College of Computer Science and Engineering, Shandong University of Science and Technology\\
$^2$School of Artificial Intelligence, Nanjing University of Information Science and Technology\\
$^3$School of Information Engineering, Guangdong University of Technology\\
$^4$School of Electronic and Information Engineering, Harbin Institute of Technology\\
{\tt\small scchu0803@gmail.com, douzhichao2021@163.com, jengshyangpan@gmail.com,}\\
{\tt\small wswweiwei@126.com, lijunbao@hit.edu.cn}
}

\begin{document}
\maketitle
\begin{abstract}
Transformer-based methods have demonstrated excellent performance on super-resolution visual tasks, surpassing conventional convolutional neural networks. However, existing work typically restricts self-attention computation to non-overlapping windows to save computational costs. This means that Transformer-based networks can only use input information from a limited spatial range. Therefore, a novel Hybrid Multi-Axis Aggregation network (HMA) is proposed in this paper to exploit feature potential information better. HMA is constructed by stacking Residual Hybrid Transformer Blocks(RHTB) and Grid Attention Blocks(GAB). On the one side, RHTB combines channel attention and self-attention to enhance non-local feature fusion and produce more attractive visual results. Conversely, GAB is used in cross-domain information interaction to jointly model similar features and obtain a larger perceptual field. For the super-resolution task in the training phase, a novel pre-training method is designed to enhance the model representation capabilities further and validate the proposed model's effectiveness through many experiments. The experimental results show that HMA outperforms the state-of-the-art methods on the benchmark dataset. We provide code and models at \href{https://github.com/korouuuuu/HMA}{https://github.com/korouuuuu/HMA}.
\end{abstract} 
\section{Introduction}
\label{sec:intro}

\begin{figure}[t]
\centering
\includegraphics[width=1.0\columnwidth]{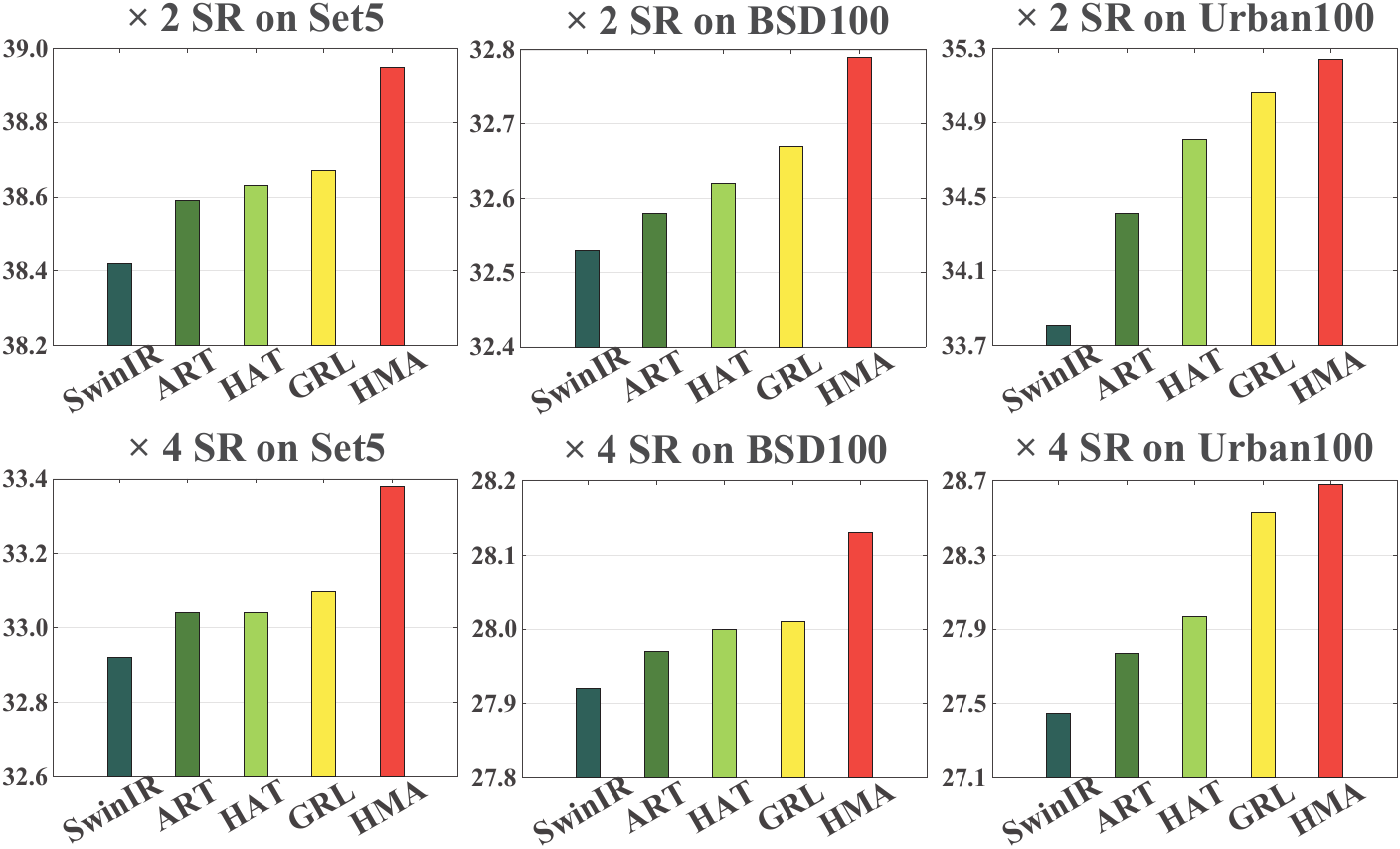}
\caption{The performance of the proposed HMA is compared with the state-of-the-art SwinIR, ART, HAT, and GRL methods in terms of PSNR(dB). Our method outperforms the state-of-the-art methods by 0.1dB$\sim$1.4dB.}
\label{fig1}
\end{figure}

Natural images have different features, such as multi-scale pattern repetition, same-scale texture similarity, and structural similarity~\cite{5995401}. Deep neural networks can exploit these properties for image reconstruction. However, it cannot capture the complex dependencies between distant elements due to the limitations of CNN's fixed local receptive field and parameter sharing mechanism, thus limiting its ability to model long-range dependencies~\cite{NIPS2016_c8067ad1}. Recent research has introduced the self-attention mechanism to computer vision~\cite{SwinIR,Lu_2022_CVPR}. Researchers have used the long-range dependency modeling capability and multi-scale processing advantages in the self-attention mechanism to enhance the joint modeling of different hierarchical structures in images.

Although Transformer-based methods have been successfully applied to image restoration tasks, there are still some things that could be improved. Existing window-based Transformer networks restrict the self-attention computation to a dense area. This strategy obviously leads to a limited receptive field and does not fully utilize the feature information from the original image. For the purpose of generating images with more realistic details, researchers consider using GAN networks or inputting the reference information to provide additional feature information~\cite{10225288,9829280,refsr}. However, the network may generate unreasonable results if the input additional feature information does not match.

In order to overcome the above problems, we propose a hybrid multiaxial aggregation network called HMA in this paper. HMA combines channel attention and self-attention, which utilizes channel attention's global information perception capability to compensate for self-attention's shortcomings. In addition, we introduce a grid attention block to achieve the modeling across distances in images. Meanwhile, to further excite the potential performance of the model, we customize a pre-training strategy for the super-resolution task. Benefiting from these designs, as shown in \cref{fig1}, our proposed method can effectively improve the model performance (0.1dB$\sim$1.4dB). The main contributions of this paper are summarised as follows:

\begin{itemize}
\item{We propose a novel Hybrid Multi-axis Aggregation network (HMA). The HMA comprises Residual Hybrid Transformer Blocks (RHTB) and Grid Attention Blocks (GAB), aiming to consider both local and global receptive fields. GAB models similar features at different image scales to achieve better reconstruction.}
\item{We further propose a pre-training strategy for super-resolution tasks that can effectively improve the model's performance using a small training cost.}
\item{Through a series of comprehensive experiments, our findings substantiate that HMA attains a state-of-the-art performance across various test datasets.}
\end{itemize} 
\section{Related Works}
\subsection{CNN-Based SISR}

CNN-based SISR methods have made significant progress in recovering image texture details. SRCNN~\cite{srcnn} solved the super-resolution task for the first time using CNNs. Subsequently, in order to enhance the network learning ability, VDSR~\cite{Kim_2016_CVPR} introduced the residual learning idea, which effectively solved the problem of gradient vanishing in deep network training. In SRGAN~\cite{Ledig_2017_CVPR}, Christian Ledig et al. proposed to use generative adversarial networks to optimize the process of generating super-resolution images. The generator of SRGAN learns the mapping from low-resolution images to high-resolution images and improves the quality of the generated images by adversarial training. ESRGAN~\cite{Wang_2018_ECCV_Workshops} introduces Residual in Residual Dense Block (RRDB) as the basic network unit and reduces the perceptual loss by using features before activation so that the images generated by EARGAN~\cite{Wang_2018_ECCV_Workshops} have a more realistic natural texture. In addition, new network architectures are constantly being proposed by researchers to recover more realistic super-resolution image details~\cite{Bilecen_2023_CVPR,SWIN2SR,Zamfir_2023_CVPR}.

\subsection{Transformer-Based SISR}

In recent years, Transformer-based SISR has become an emerging research direction in super-resolution, which utilizes the Transformer architecture to achieve image mapping from low to high resolution. Among them, the Swin Transformer-based SwinIR~\cite{SwinIR} model achieves the best performance beyond CNN-based on image restoration tasks. In order to further investigate the effect of pre-training on its internal representation, Chen et al. proposed a novel Hybrid Attention Transformer (HAT)~\cite{Chen_2023_CVPR}. The HAT introduces overlapping cross-attention blocks to enhance the interactions between neighboring windows' features, thus aggregating the cross-window information better. Our proposed HMA network learns similar feature representations through grid multiplexed self-attention and combines it with channel attention to enhance non-local feature fusion. Therefore, our method can provide additional support for image restoration through similar features in the original image.

\subsection{Self-similarity based image restoration}

Natural images usually have similar features in different hierarchies, and many SISR methods based on CNN have achieved remarkable results by exploring self-similarity~\cite{Huang_2015_CVPR,9992208,Mei_2020_CVPR}. In order to reduce the computational complexity, the computation of self-similarity is usually restricted to local areas. The researchers also proposed to extend the search space by geometric transformations to increase the global feature interactions~\cite{selfsimilarity}. In Transformer-based SISR, the computational complexity of non-local self-attention increases quadratically with the growth of image size. Recent studies have proposed using sparse global self-attention to reduce the complexity~\cite{zhang2022accurate}. Sparse global self-attention allows more feature interactions while reducing computational complexity. The proposed GAB adopts the idea of sparse self-attention to increase global feature interactions while balancing the computational complexity. Our method allows joint modeling using similar features to generate better reconstructed images. 
\section{Motivation}

\begin{figure}[t]
\centering
\includegraphics[width=1.0\columnwidth]{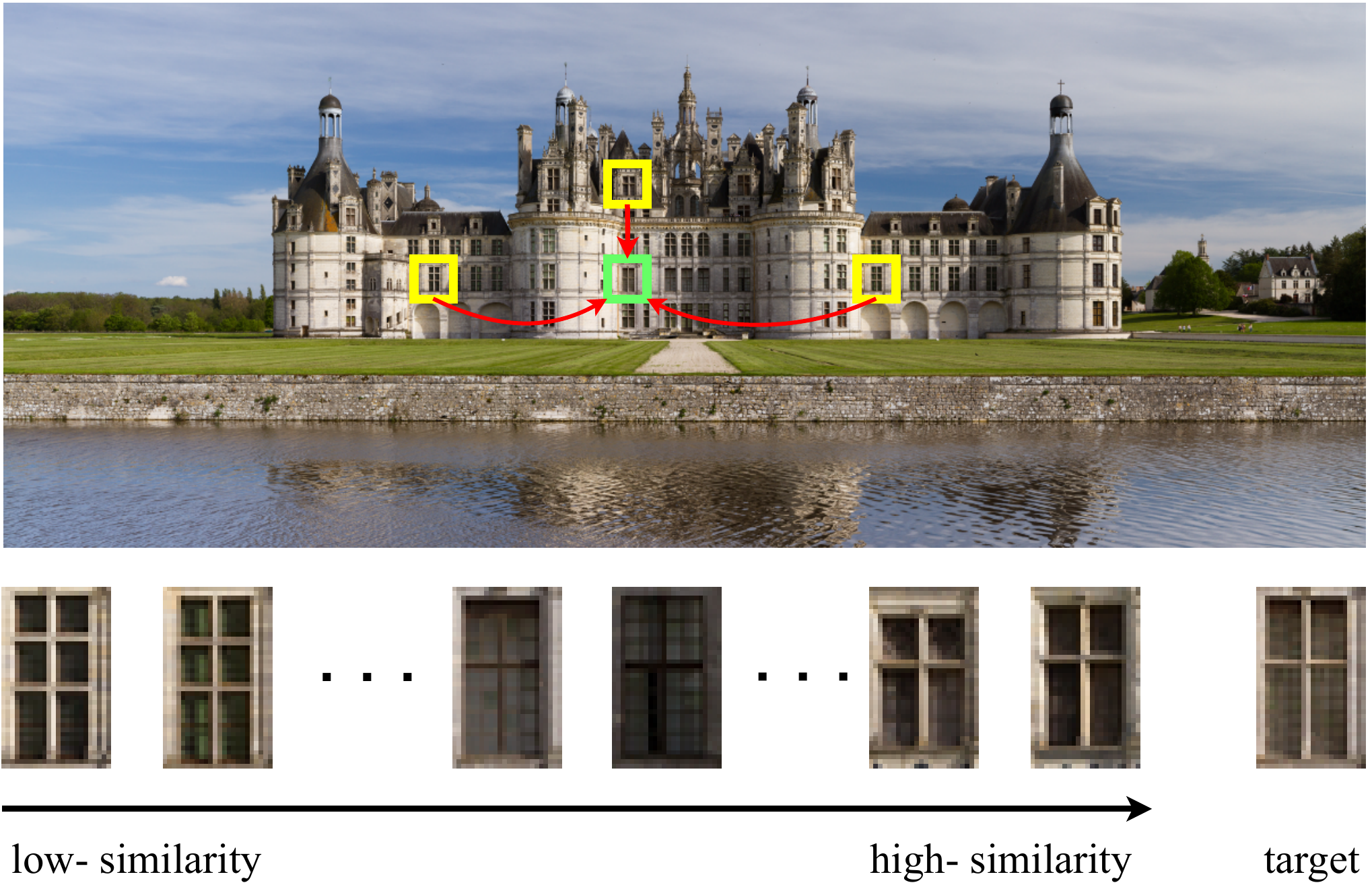}
\caption{Example of image similarity based on non-local textures. Image from DIV2K:0830.}
\label{fig2}
\end{figure}

Image self-similarity is vital in image processing, computer vision, and pattern recognition. Image self-similarity is usually characterized by multi-scale and geometric transformation invariance. Image self-similarity can be local or global. Local self-similarity means that one area of an image is similar to another, and global self-similarity means that there is self-similarity between multiple areas within the whole image. \cref{fig2} shows that texture units may be repeated at regular intervals. Similarity modeling of features at different locations (\textit{e.g.}, yellow rectangle) in the input image can provide a reference for image reconstruction in the green rectangle when recovering the features in the green rectangle. Image self-similarity has been explored with satisfactory performance in classical super-resolution algorithms.

\begin{figure}[t]
\centering
\includegraphics[width=0.7\columnwidth]{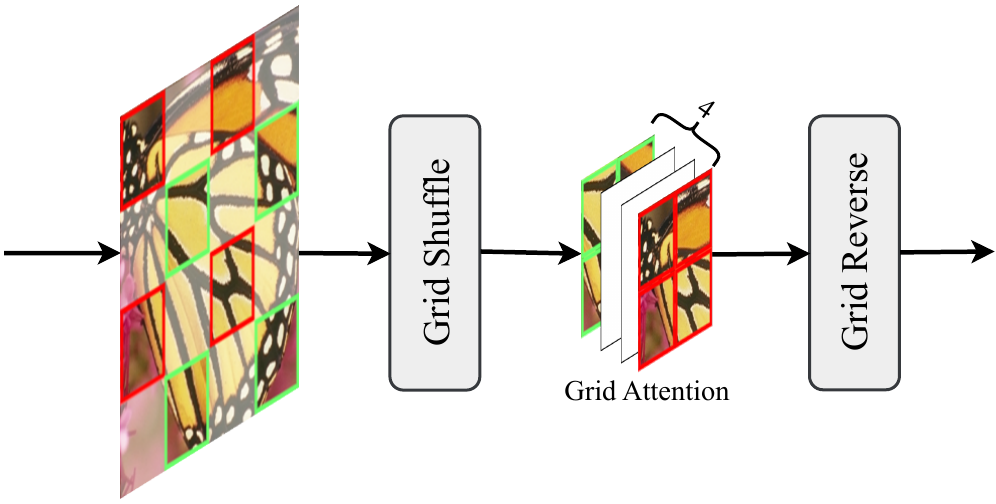}
\caption{Grid Attention Strategies. We divide the feature map into sparse areas at specific intervals ($K=4$) and then compute the self-attention within each set of sparse areas.}
\label{fig3}
\end{figure}

Swin Transformer~\cite{Liu_2021_ICCV} employs cross-window connectivity and multi-head attention mechanisms to deal with the long-range dependency modeling problem. However, Swin Transformer can only use a limited range of pixels when dealing with the SR task and cannot effectively use image self-similarity to enhance the reconstruction effect. For the purpose of increasing the range of pixels utilized by the Swin Transformer, we try to enhance the long-range dependency modeling capability of the Swin Transformer with sparse attention. As shown in \cref{fig3}, we suggest adding grid attention to increase the interaction between patches. The feature map is divided into $K^2$ groups according to the interval size $K$, and each group contains $\frac{H}{K}\times \frac{W}{K}$ Patches. After the grid shuffle, we can get the feature $F_G\in \mathbb{R}^{\frac{H}{K}\times \frac{W}{K}\times C}$ and compute the self-attention in each group.

Not all areas in a natural image have similarity relationships. In order to avoid the non-similar features from damaging the original features, we introduce the global feature-based interaction feature $G\in \mathbb{R}^{\frac{H}{K}\times \frac{W}{K}\times \frac{C}{2}}$ and the window-based self-attention mechanism ((S)W-MSA) to capture the similarity relationship of the whole image while modeling the similar features by Grid Multihead Self-Attention (Grid-MSA). The detailed computational procedure is described in \cref{sec4.c}.

\begin{figure}[t]
\centering
\includegraphics[width=1.0\columnwidth]{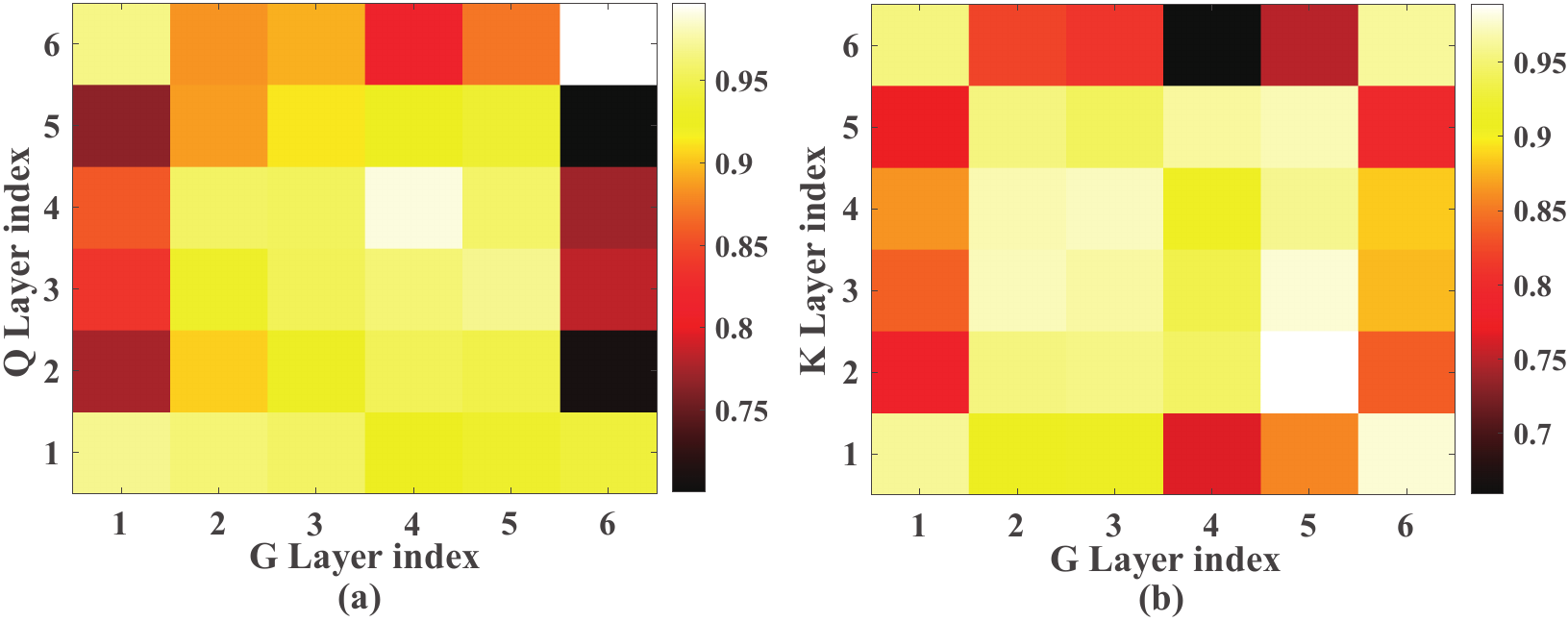}
\caption{(a) CKA similarity between all G and Q in the $\times$2 SR model. (b) CKA similarity between all G and K in the $\times$2 SR model.}
\label{fig4}
\end{figure}

To make Grid-MSA work better, we must ensure the similarity between interaction features and query/key structure. Therefore, we introduce centered kernel alignment (CKA)~\cite{pmlr-v97-kornblith19a} to study the similarity between features. It can be observed that the CKA similarity maps in \cref{fig4} presents a diagonal structure, \textit{i.e.}, there is a close structural similarity between the interaction features and the query/keyword in the same layer (CKA$>$0.9). Therefore, interaction features can be a medium for query/key interaction with global features in Grid-MSA. With the benefit of these designs, our network is able to reconstruct the image taking full advantage of the pixel information in the input image. 
\section{Proposed Method}

\begin{figure*}[t]
\centering
\includegraphics[width=1.0\textwidth]{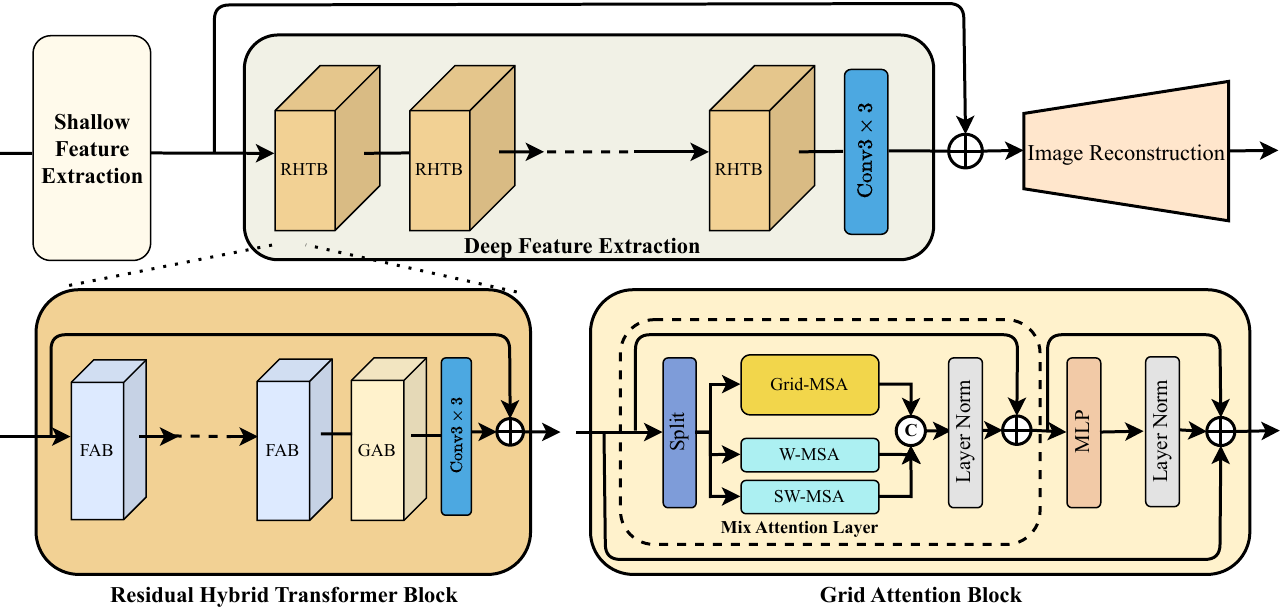}
\caption{The overall architecture of HMA and the structure of RHTB and GAB.}
\label{fig5}
\end{figure*}

As shown in \cref{fig5}, HMA consists of three parts: shallow feature extraction, deep feature extraction, and image reconstruction. Among them, RHTB is a stacked combination of multiple Fused Attention Blocks (FAB) and GAB. The RHTB is constructed by residual in residual structure. We will introduce these methods in detail in the following sections.

\subsection{Overall Architecture}

For a given low-resolution (LR) input $I_{LR}\in \mathbb{R}^{H\times W\times C_{in}}$ ($H$, $W$, and $C_{in}$ are the height, width, and number of input channels of the input image, respectively), we first extract the shallow features of the $I_{LR}$ using a convolutional layer that maps the $I_{LR}$ to high-dimensional features $F_0\in \mathbb{R}^{H\times W\times C}$ :

\begin{equation}
\label{eq.1}
F_0=H_{Conv}(I_{LR}),
\end{equation}
where $H_{Conv} (\cdot)$ denotes the convolutional layer and $C$ denotes the number of channels of the intermediate layer features. Subsequently, we input $F_0$ into $H_{DF} (\cdot)$, a deep feature extraction group consisting of M RHTBs and a 3 $\times$ 3 convolution. Each RHTB consists of a stack of N FABs, a GAB, and a convolutional layer with residual connections. Then, we fuse the deep features $F_D \in \mathbb{R}^{H \times W \times C}$ with $F_0$ by element-by-element summation to obtain $F_{REC}$. Finally, we reconstruct $F_{REC}$ into a high-resolution image $I_{HR}$:

\begin{equation}
\label{eq.2}
I_{HR} = H_{REC}(H_{DF}(F_0)+F_0),
\end{equation}
where $H_{REC} (\cdot)$ denotes the reconstruction module.

\subsection{Fused Attention Block (FAB)}

\begin{figure}[t]
\centering
\includegraphics[width=1.0\columnwidth]{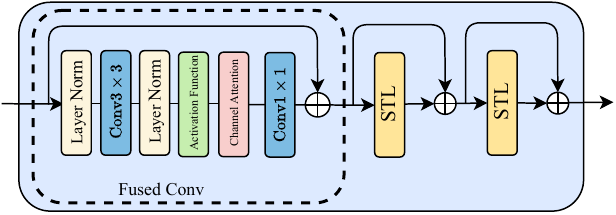}
\caption{The architecture of FAB.}
\label{fig6}
\end{figure}

Many studies have shown that adding appropriate convolution in the Transformer can further improve network trainability~\cite{9779501,EdgeNeXt,Yoo_2023_WACV}. Therefore, we insert a convolutional layer before the Swin Transformer Layer (STL) to enhance the network learning capability. As shown in \cref{fig6}, we insert the Fused Conv module ($H_{Fuse} (\cdot)$) with inverted bottlenecks and squeezed excitations before the STL to achieve enhanced global information fusion. Note that we use Layer Norm instead of Batch Norm in Fused Conv to avoid the impact on the contrast and color of the image. The computational procedure of Fused Conv is:

\begin{equation}
\label{eq.3}
F_{Fuse} = H_{Fuse}(F_{F_{in}}) + F_{F_{in}},
\end{equation}
where $F_{F_{in}}$ represents the input features, and $F_{Fuse}$ represents the features output from the Fused Conv block. Then, we add two successive STL after Fused Conv. In the STL, we follow the classical design in SWinIR, including Window-based self-attention (W-MSA) and Shifted Window-based self-attention (SW-MSA), and Layer Norm. The computation of the STL is as follows:

\begin{equation}
\label{eq.4}
F_N=(S)W-MSA(LN(F_{W_{in}}))+F_{W_{in}},
\end{equation}

\begin{equation}
\label{eq.5}
F_{out}=MLP(LN(F_N))+F_N,
\end{equation}

where $F_{W_{in}}$, $F_N$, and $F_{out}$ indicate the input features, the intermediate features, and the output of the STL, respectively, and MLP denotes the multilayer perceptron. We split the feature map uniformly into $\frac{H\times W}{M^2}$ windows in a non-overlapping manner for efficient modeling. Each window contains M $\times$ M Patch. The self-attention of a local window is calculated as follows:

\begin{equation}
\label{eq.6}
\mathit{Attention}(Q,K,V)=SoftMax(\frac{QK^T}{\sqrt{d}}+B)V,
\end{equation}
where $Q$, $K$, $V\in \mathbb{R}^{M^2\times d}$ are obtained by the linear transformation of the given input feature $F_W\in \mathbb{R}^{M^2\times C}$. The $d$ and $B$ represent the dimension and relative position encoding of the query/key, respectively.

As shown in \cref{fig6}, Fused Conv expands the channel using a convolutional kernel of size 3 with a default expansion rate of 6. At the same time, a squeeze-excitation (SE) layer with a shrink rate of 0.5 is used in the channel attention layer. Finally, a convolutional kernel of size 1 is used to recover the channel.

\subsection{Grid Attention Block(GAB)}
\label{sec4.c}

\begin{figure}[t]
\centering
\includegraphics[width=1.0\columnwidth]{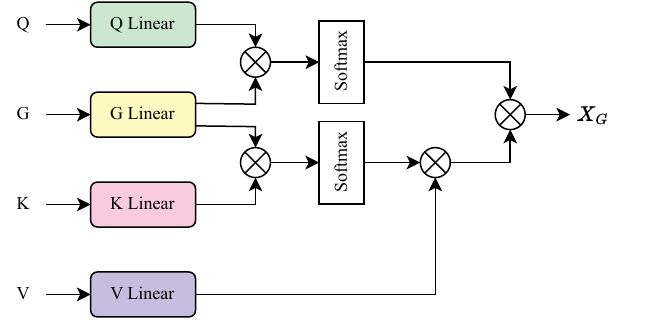}
\caption{The computational flowchart of Grid Attention.}
\label{fig7}
\end{figure}

We introduce GAB to model cross-area similarity for enhanced image reconstruction. The GAB consists of a Mix Attention Layer (MAL) and an MLP layer. Regarding the MAL, we first split the input feature $F_{in}$ into two parts by channel: $F_G\in \mathbb{R}^{H\times W\times \frac{C}{2}}$ and $F_W\in \mathbb{R}^{H\times W\times \frac{C}{2}}$. Subsequently, we split $F_W$ into two parts by channel again and input them into W-MSA and SW-MSA, respectively. Meanwhile, $F_G$ is input into Grid-MSA.The computation process of MAL is as follows:

\begin{equation}
\label{eq.7}
X_{W_1}=W-MSA(F_{W_1}),
\end{equation}

\begin{equation}
\label{eq.8}
X_{W_2}=SW-MSA(F_{W_2}),
\end{equation}

\begin{equation}
\label{eq.9}
X_G=Grid-MSA(F_G),
\end{equation}

\begin{equation}
\label{eq.10}
X_{MAL}=LN(Cat(X_{W_1},X_{W_2},X_G))+F_{in},
\end{equation}
where $X_{W_1}$, $X_{W_2}$, and $X_G$ are the output features of W-MSA, SW-MSA, and Grid-MSA, respectively. It should be noted that we adopt the post-norm method in GAB to enhance the network training stability. For a given input feature $F_{in}$, the computation process of GAB is:

\begin{equation}
\label{eq.11}
F_M=LN(MAL(F_{in})+F_{in},
\end{equation}

\begin{equation}
\label{eq.12}
F_{out}=LN(MLP(F_M))+F_M,
\end{equation}

It is shown in \cref{fig7} that the Q, K, and V are obtained from the input feature $F_G$ after grid shuffle when Grid-MSA is used. $G\in \mathbb{R}^{H\times W\times \frac{C}{2}}$ is obtained from the linear transformation of the input feature $F_{in}$ after grid shuffle. For Grid-MSA, the self-attention is calculated as follows:

\begin{equation}
\label{eq.13}
\hat{X}=SoftMax(\frac{GK^T}{d}+B)V,
\end{equation}

\begin{equation}
\label{eq.14}
\mathit{Attention}(Q,G,\hat{X})=SoftMax(\frac{QG^T}{d}+B)\hat{X},
\end{equation}
where $\hat{X}$ is the intermediate feature obtained by computing the self-attention from $G$, $K$, and $V$.

\subsection{Pre-training strategy}

Pre-training plays a crucial role in many visual tasks~\cite{MultiMAE,pmlr-v202-wang23e}. Recent studies have shown that pre-training can also capture significant gains in low-level visual tasks. IPT~\cite{Chen_2021_CVPR} handles different visual tasks by sharing the Transformer module with different head and tail structures. EDT~\cite{li2021efficient} improves the performance of the target task by multi-task pre-training. HAT~\cite{Chen_2023_CVPR} pre-trains the super-resolution task using a larger dataset directly on the same task. Instead, we propose a pre-training method more suitable for super-resolution tasks, \textit{i.e.}, increasing the gain of pre-training by sharing model parameters among pre-trained models with different degradation levels. We first train a $\times$2 model as the initial parameter seed when pre-training on the ImageNet dataset and then use it as the initialization parameter for the $\times$3 model. Then, train the final $\times$2 and $\times$4 models using the trained $\times$3 model as the initialization parameters of the $\times$2 and $\times$4 models. After the pre-training, the $\times$2, $\times$3, and $\times$4 models are fine-tuned on the DF2K dataset. The proposed strategy can bring more performance improvement, although it pays an extra training cost (training a $\times$2 model). 
\section{Experiments}

\subsection{Experimental Setup}

We use DF2K dataset (DIV2K~\cite{Lim_2017_CVPR_Workshops} dataset merged with Flicker~\cite{Timofte_2017_CVPR_Workshops} dataset) as the training set. Meanwhile, we use ImageNet~\cite{5206848} as the pre-training dataset. For the structure of HMA, the number of RHTB and FAB is set to 6, the window size is set to 16, the number of channels is set to 180, and the number of attentional heads is set to 6. The number of attentional heads is 3 and 2 for Grid-MSA and (S)W-MSA in GAB, respectively. We evaluate on the Set5~\cite{set5}, Set14~\cite{set14}, BSD100~\cite{bsd100}, Urban100~\cite{Huang_2015_CVPR}, and Manga109~\cite{matsui2017sketch} datasets. Both PSNR and SSIM evaluations are computed on the Y channel.

\subsection{Training Details}

Low-resolution images are generated by down-sampling using bicubic interpolation in MATLAB. We cropped the dataset into 64$\times$64 patches for training. Furthermore, we employed horizontal flipping and random rotation for data augmentation. The training batch size is set to 32. During pre-training with ImageNet~\cite{5206848}, the total number of training iterations is set to 800K (1K represents 1000 iterations), the learning rate was initialized to $2\times 10^{-4}$ and halved at [300K, 500K, 650K, 700K, 750K]. We optimized the model using the Adam optimizer (with $\beta_1$=0.9 and $\beta_2$=0.99). Subsequently, we fine-tuned the model on the DF2K dataset. The total number of training iterations is set to 250K, and the initial learning rate was set to $5\times 10^{-6}$ and halved at [125K, 200K, 230K, 240K].

\subsection{Ablation Study}
\subsubsection{Effectiveness of Fused Conv and GAB}
\begin{table}
  \begin{center}

  \resizebox{0.9\columnwidth}{!}{
    \begin{tabular}{|c|cccc|}
    \hline
          & \multicolumn{4}{c|}{Baseline}\\
    \hline
    Fused Conv & \usym{2717}     & \usym{2717}     & \usym{2713}     & \usym{2713} \\
    GAB   & \usym{2717}     & \usym{2713}     & \usym{2717}     & \usym{2713}\\
    \hline
    PSNR/SSIM  & 27.49/0.8271 & 28.30/0.8370 & 28.37/0.8375 & 28.42/0.8450 \\
    \hline
    \end{tabular}%
    }
    \end{center}
    \caption{Ablation study on the proposed Fused Conv and GAB.}\label{tab2}%
\end{table}%

\begin{table}
  \centering

    \begin{tabular}{|c|cccc|}
    \hline
    expansion rate & 2     & 4     & 6     & 8\\
    \hline
    PSNR  & 28.30 & 28.34 & 28.37 & 28.39\\
    \hline
    \end{tabular}%
    \caption{Ablation study on expansion rate of Fused Conv.}\label{tab3}%
\end{table}%

\begin{table}
  \centering

    \begin{tabular}{|c|cccc|}
    \hline
    shrink rate & 2     & 4     & 6     & 8 \\
    \hline
    PSNR  & 27.39 & 28.37 & 28.32 & 28.28 \\
    \hline
    \end{tabular}%
    \caption{Ablation study on shrink rate of Fused Conv.}\label{tab4}%
\end{table}%

\begin{table*}[!ht]
  \centering

    \begin{tabular}{|c|c|c|c|c|c|c|c|c|c|c|c|}
    \hline
    \multirow{2}{*}{Method} & \multirow{2}{*}{Scale} & \multicolumn{2}{c|}{Set5\cite{set5}} & \multicolumn{2}{c|}{Set14\cite{set14}} & \multicolumn{2}{c|}{BSD100\cite{bsd100}} & \multicolumn{2}{c|}{Urban100\cite{Huang_2015_CVPR}} & \multicolumn{2}{c|}{Manga109\cite{matsui2017sketch}}\\
\cline{3-12}          &       & PSNR  & SSIM  & PSNR  & SSIM  & PSNR  & SSIM  & PSNR  & SSIM  & PSNR  & SSIM \\
    \hline
    EDSR\cite{Lim_2017_CVPR_Workshops}  & $\times$2    & 38.11 & 0.9602 & 33.92 & 0.9195 & 32.32 & 0.9013 & 32.93 & 0.9351 & 39.10  & 0.9773 \\
    RCAN\cite{Zhang_2018_ECCV}  & $\times$2    & 38.27 & 0.9614 & 34.12 & 0.9216 & 32.41 & 0.9027 & 33.34 & 0.9384 & 39.44 & 0.9786 \\
    SAN\cite{san}   & $\times$2    & 38.31 & 0.9620 & 34.07 & 0.9213 & 32.42 & 0.9028 & 33.10  & 0.9370 & 39.32 & 0.9792 \\
    IGNN\cite{ignn}  & $\times$2    & 38.24 & 0.9613 & 34.07 & 0.9217 & 32.41 & 0.9025 & 33.23 & 0.9383 & 39.35 & 0.9786 \\
    NLSA\cite{nlsa}  & $\times$2    & 38.34 & 0.9618 & 34.08 & 0.9231 & 32.43 & 0.9027 & 33.42 & 0.9394 & 39.59 & 0.9789 \\
    IPT\cite{Chen_2021_CVPR}   & $\times$2    & 38.37 & -     & 34.43 & -     & 32.48 & -     & 33.76 & -     & -     & - \\
    SwinIR\cite{SwinIR} & $\times$2    & 38.42 & 0.9623 & 34.46 & 0.9250 & 32.53 & 0.9041 & 33.81 & 0.9427 & 39.92 & 0.9797 \\
    ESRT\cite{lu2022transformer} & $\times$2    & -     & -     & -     & -     & -     & -     & -     & -     & -     & - \\
    SRFormer\cite{zhou2023srformer} & $\times$2    & 38.51 & 0.9627 & 34.44 & 0.9253 & 32.57 & 0.9046 & 34.09 & 0.9449 & 40.07 & 0.9802 \\
    EDT\cite{li2021efficient}   & $\times$2    & 38.45 & 0.9624 & 34.57 & 0.9258 & 32.52 & 0.9041 & 33.80  & 0.9425 & 39.93 & 0.9800 \\
    HAT\cite{Chen_2023_CVPR}   & $\times$2    & 38.63 & 0.9630 & 34.86 & 0.9274 & 32.62 & 0.9053 & 34.45 & 0.9466 & 40.26 & 0.9809 \\
    GRL\cite{Li_2023_CVPR}   & $\times$2    & 38.67 & \textcolor{blue}{0.9647} & 35.08 & \textcolor{red}{0.9303} & \textcolor{green}{32.68} & \textcolor{red}{0.9087} & \textcolor{green}{35.06} & \textcolor{blue}{0.9505} & 40.67 & 0.9818 \\
    HMA(ours)   & $\times$2    & \textcolor{green}{38.79} & 0.9641 & \textcolor{green}{35.11} & 0.9286 & 32.67 & 0.9061 & 34.85 & \textcolor{green}{0.9493} & \textcolor{green}{40.73} & \textcolor{green}{0.9824} \\
    \hdashline
    HAT-L$^\dag$\cite{Chen_2023_CVPR} & $\times$2    & \textcolor{blue}{38.91} & \textcolor{green}{0.9646} & \textcolor{blue}{35.29} & \textcolor{green}{0.9293} & \textcolor{blue}{32.74} & \textcolor{green}{0.9066} & \textcolor{blue}{35.09} & \textcolor{blue}{0.9505} & \textcolor{blue}{41.01} & \textcolor{blue}{0.9831} \\
    HMA$^\dag$(ours)   & $\times$2    & \textcolor{red}{38.95} & \textcolor{red}{0.9649} & \textcolor{red}{35.33} & \textcolor{blue}{0.9297} & \textcolor{red}{32.79} & \textcolor{blue}{0.9071} & \textcolor{red}{35.24} & \textcolor{red}{0.9513} & \textcolor{red}{41.13} & \textcolor{red}{0.9836}\\
    \hline
    EDSR\cite{Lim_2017_CVPR_Workshops}  & $\times$3    & 34.65 & 0.928 & 30.52 & 0.8462 & 29.25 & 0.8093 & 28.80  & 0.8653 & 34.17 & 0.9476\\
    RCAN\cite{Zhang_2018_ECCV}  & $\times$3    & 34.74 & 0.9299 & 30.65 & 0.8482 & 29.32 & 0.8111 & 29.09 & 0.8702 & 34.44 & 0.9499 \\
    SAN\cite{san}   & $\times$3    & 34.75 & 0.9300  & 30.59 & 0.8476 & 29.33 & 0.8112 & 28.93 & 0.8671 & 34.30  & 0.9494 \\
    IGNN\cite{ignn}  & $\times$3    & 34.72 & 0.9298 & 30.66 & 0.8484 & 29.31 & 0.8105 & 29.03 & 0.8696 & 34.39 & 0.9496 \\
    NLSA\cite{nlsa}  & $\times$3    & 34.85 & 0.9306 & 30.70  & 0.8485 & 29.34 & 0.8117 & 29.25 & 0.8726 & 34.57 & 0.9508 \\
    IPT\cite{Chen_2021_CVPR}   & $\times$3    & 34.81 & -     & 30.85 & -     & 29.38 & -     & 29.49 & -     & -     & - \\
    SwinIR\cite{SwinIR} & $\times$3    & 34.97 & 0.9318 & 30.93 & 0.8534 & 29.46 & 0.8145 & 29.75 & 0.8826 & 35.12 & 0.9537 \\
    ESRT\cite{lu2022transformer} & $\times$3    & 34.42 & 0.9268 & 30.43 & 0.8433 & 29.15 & 0.8063 & 28.46 & 0.8574 & 33.95 & 0.9455 \\
    SRFormer\cite{zhou2023srformer} & $\times$3    & 35.02 & 0.9323 & 30.94 & 0.8540 & 29.48 & 0.8156 & 30.04 & 0.8865 & 35.26 & 0.9543 \\
    EDT\cite{li2021efficient}   & $\times$3    & 34.97 & 0.9316 & 30.89 & 0.8527 & 29.44 & 0.8142 & 29.72 & 0.8814 & 35.13 & 0.9534 \\
    HAT\cite{Chen_2023_CVPR}   & $\times$3    & 35.07 & 0.9329 & 31.08 & 0.8555 & 29.54 & 0.8167 & 30.23 & 0.8896 & 35.53 & 0.9552 \\
    GRL\cite{Li_2023_CVPR}   & $\times$3    & -     & -     & -     & -     & -     & -     & -     & -     & -     & - \\
    HMA(ours)   & $\times$3    & \textcolor{green}{35.22} & \textcolor{green}{0.9336} & \textcolor{green}{31.28} & \textcolor{green}{0.8570} & \textcolor{green}{29.59} & \textcolor{green}{0.8682} & \textcolor{green}{30.65} & \textcolor{green}{0.8944} & \textcolor{green}{35.82} & \textcolor{green}{0.9567} \\
    \hdashline
    HAT-L$^\dag$\cite{Chen_2023_CVPR} & $\times$3    & \textcolor{blue}{35.28} & \textcolor{blue}{0.9345} & \textcolor{blue}{31.47} & \textcolor{blue}{0.8584} & \textcolor{blue}{29.63} & \textcolor{blue}{0.8191} & \textcolor{blue}{30.92} & \textcolor{blue}{0.8981} & \textcolor{blue}{36.02} & \textcolor{blue}{0.9576} \\
    HMA$^\dag$(ours)   & $\times$3    & \textcolor{red}{35.35} & \textcolor{red}{0.9347} & \textcolor{red}{31.47} & \textcolor{red}{0.8585} & \textcolor{red}{29.66} & \textcolor{red}{0.8196} & \textcolor{red}{31.00} & \textcolor{red}{0.8984} & \textcolor{red}{36.10} & \textcolor{red}{0.9580}\\
    \hline
    EDSR\cite{Lim_2017_CVPR_Workshops}  & $\times$4    & 32.46 & 0.8968 & 28.80  & 0.7876 & 27.71 & 0.7420 & 26.64 & 0.8033 & 31.02 & 0.9148\\
    RCAN\cite{Zhang_2018_ECCV}  & $\times$4    & 32.63 & 0.9002 & 28.87 & 0.7889 & 27.77 & 0.7436 & 26.82 & 0.8087 & 31.22 & 0.9173 \\
    SAN\cite{san}   & $\times$4    & 32.64 & 0.9003 & 28.92 & 0.7888 & 27.78 & 0.7436 & 26.79 & 0.8068 & 31.18 & 0.9169 \\
    IGNN\cite{ignn}  & $\times$4    & 32.57 & 0.8998 & 28.85 & 0.7891 & 27.77 & 0.7434 & 26.84 & 0.8090 & 31.28 & 0.9182 \\
    NLSA\cite{nlsa}  & $\times$4    & 32.59 & 0.9000   & 28.87 & 0.7891 & 27.78 & 0.7444 & 26.96 & 0.8109 & 31.27 & 0.9184 \\
    IPT\cite{Chen_2021_CVPR}   & $\times$4    & 32.64 & -     & 29.01 & -     & 27.82 & -     & 27.26 & -     & -     & - \\
    SwinIR\cite{SwinIR} & $\times$4    & 32.92 & 0.9044 & 29.09 & 0.7950 & 27.92 & 0.7489 & 27.45 & 0.8254 & 32.03 & 0.9260 \\
    ESRT\cite{lu2022transformer} & $\times$4    & 32.19 & 0.8947 & 28.69 & 0.7833 & 27.69 & 0.7379 & 26.39 & 0.7962 & 30.75 & 0.9100 \\
    SRFormer\cite{zhou2023srformer} & $\times$4    & 32.93 & 0.9041 &  29.08 &  0.7953 & 27.94 & 0.7502 & 27.68 & 0.8311 & 32.21 & 0.9271 \\
    EDT\cite{li2021efficient}   & $\times$4    & 32.82 & 0.9031 & 29.09 & 0.7939 & 27.91 & 0.7483 & 27.46 & 0.8246 & 32.05 & 0.9254 \\
    HAT\cite{Chen_2023_CVPR}   & $\times$4    & 33.04 & 0.9056 & 29.23 & 0.7973 & 28.00    & 0.7517 & 27.97 & 0.8368 & 32.48 & 0.9292 \\
    GRL\cite{Li_2023_CVPR}   & $\times$4    & 33.10  & \textcolor{red}{0.9094} & \textcolor{green}{29.37} & \textcolor{red}{0.8058} & 28.01 & \textcolor{red}{0.7611} & \textcolor{green}{28.53} & \textcolor{blue}{0.8504} & 32.77 & \textcolor{green}{0.9325} \\
    HMA(ours)   & $\times$4    & \textcolor{green}{33.15} & 0.9060 & 29.32 & 0.7996 & \textcolor{green}{28.05} & 0.7530 & 28.42 & 0.8450 & \textcolor{green}{32.97} & 0.9320 \\
    \hdashline
    HAT-L$^\dag$\cite{Chen_2023_CVPR} & $\times$4    & \textcolor{blue}{33.30}  & \textcolor{green}{0.9083} & \textcolor{blue}{29.47} & \textcolor{green}{0.8015} & \textcolor{blue}{28.09} & \textcolor{green}{0.7551} & \textcolor{blue}{28.60}  & \textcolor{green}{0.8498} & \textcolor{blue}{33.09} & \textcolor{blue}{0.9335} \\
    HMA$^\dag$(ours)   & $\times$4    & \textcolor{red}{33.38} & \textcolor{blue}{0.9089} & \textcolor{red}{29.51} & \textcolor{blue}{0.8019} & \textcolor{red}{28.13} & \textcolor{blue}{0.7562} & \textcolor{red}{28.69} & \textcolor{red}{0.8512} & \textcolor{red}{33.19} & \textcolor{red}{0.9344}\\
    \hline
    \end{tabular}%
    \caption{Quantitative comparison (PSNR/SSIM) with state-of-the-art methods on benchmark dataset. The top three results are marked in \textcolor{red}{red}, \textcolor{blue}{blue} and \textcolor{green}{green}., respectively. “$\dag$” indicates that methods adopt pre-training strategy.}\label{tab5}%
\end{table*}%

We experimentally demonstrate the effectiveness of Fused Conv and GAB proposed in this paper. The experiments are conducted on the Urban100~\cite{Huang_2015_CVPR} dataset to evaluate PSNR/SSIM. The evaluation report is presented in \cref{tab2}. Compared with the baseline results, the best performance is achieved when both modules are used. In contrast, the performance gains obtained when using the Fused Conv or GAB modules alone were not as good as when using them simultaneously. Although the performance of the sole use of the Fused Conv module is slightly higher than the sole use of the GAB module, the GAB module is applied for global image interaction, which can effectively improve the model SSIM value and better restore the image's texture. This means that our proposed method not only performs well on PSNR but is also excellent in restoring the image's visual effect.

\subsubsection{Effects of the expansion rate and shrink rate}

\begin{figure*}[!ht]
\centering
\includegraphics[width=1.0\textwidth]{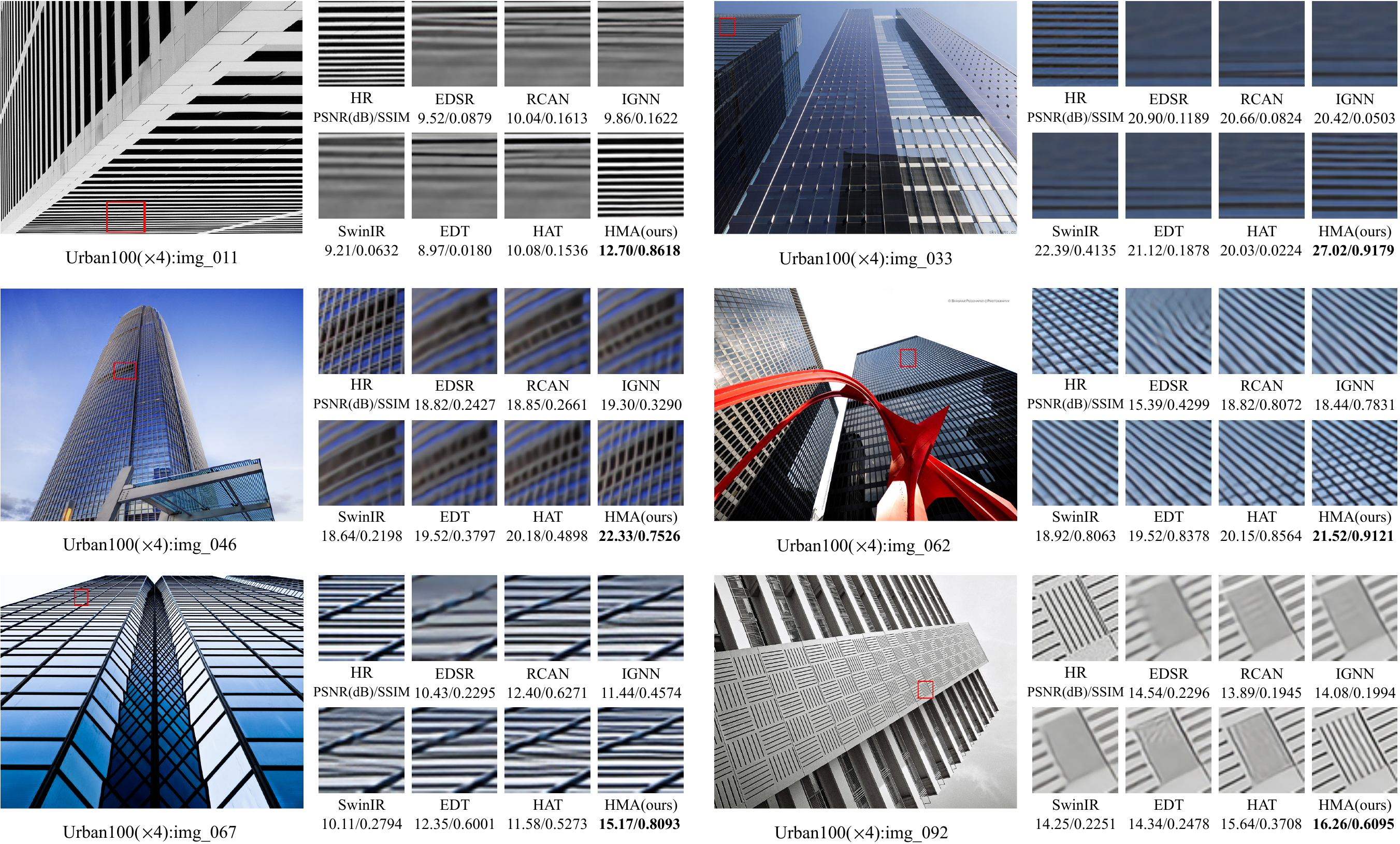}
\caption{Visual comparison on $\times$4 SR. PSNR/SSIM is calculated in patches marked with red boxes in the images.}
\label{fig9}
\end{figure*}

\cref{tab3} and \cref{tab4} show the effect of expansion and shrink rates on performance, respectively. The data in the table shows that the expansion rate is directly proportional to the performance, while the shrink rate is inversely proportional. Although the performance keeps increasing when the expansion rate increases, the number of parameters and the amount of computation increase quadratically. In order to balance the model performance and computation, we set the expansion rate to 6. Similarly, we set the shrink rate to 2 to get a model with as little computation as possible.

\subsection{Comparison with State-of-the-Art Methods}

\subsubsection{Quantitative comparison}

\cref{tab5} shows the comparative results of our method with the state-of-the-art methods on PSNR and SSIM: EDSR~\cite{Lim_2017_CVPR_Workshops}, RCAN~\cite{Zhang_2018_ECCV}, SAN~\cite{san}, IGNN~\cite{ignn}, NLSA~\cite{nlsa}, IPT~\cite{Chen_2021_CVPR}, SwinIR~\cite{SwinIR}, ESRT~\cite{lu2022transformer}, SRFoemer~\cite{zhou2023srformer} EDT~\cite{li2021efficient}, HAT~\cite{Chen_2023_CVPR}, HAT-L~\cite{Chen_2023_CVPR}, and GRL~\cite{Li_2023_CVPR}. In \cref{tab5}, it can be seen that the proposed method achieves the best performance on almost all scales on five datasets. Specifically, HMA outperforms SwinIR by 0.2dB$\sim$1.43dB on all scales. In particular, on Urban100~\cite{Huang_2015_CVPR} and MANGA109~\cite{matsui2017sketch} that contain a large number of repetitive textures, HMA improves by 0.98dB$\sim$1.43dB compared to SwinIR. It is important to note that both HAT and GRL~\cite{Li_2023_CVPR} introduce the channel attention in the model. However, both HAT~\cite{Chen_2023_CVPR} and GRL~\cite{Li_2023_CVPR} perform less well than HMA, which proves the effectiveness of our proposed method.

\subsubsection{Visual comparison}

We provide some of the visual comparison results in \cref{fig9}. The comparison results are selected from the Urban100~\cite{Huang_2015_CVPR} dataset: "img\underline{~}011", "img\underline{~}033", "img\underline{~}046", "img\underline{~}062", "img\underline{~}067" and "img\underline{~}092". In \cref{fig9}, PSNR and SSIM is calculated in patches marked with red boxes in the images. From the visual comparison, HMA can recover the image texture details better. Compared with other advanced methods, HMA recovers images with clearer edges. We can see many blurred areas in recovering image "img\underline{~}011" and image "img\underline{~}092" in other state-of-the-art methods, while HMA generates excellent visual effects. The comparison of the visual effects indicates that our proposed method also achieves a superior performance.

\subsection{NTIRE 2024 Challenge}
Our SR model also participated in NTIRE 2024 Image Super-Resolution ($\times$4)~\cite{chen2024ntire_sr} in the validation phase and testing phase. The respective results areshown in \cref{tab6}.
\begin{table}[htbp]
  \centering
  \caption{ NTIRE 2024 Challenge Results with $\times$4 SR in terms of PSNR and SSIM on validation phase and testing phase.}
    \begin{tabular}{ccc}
    \hline
          & Validation phase &  Testing phase \\
    \hline
    PSNR  & 31.44 & 31.18 \\
    SSIM  & 0.85  & 0.86 \\
    \hline
    \end{tabular}\label{tab6}
\end{table}%
\section{Conclusion}

This study proposes a Hybrid Multi-Axis Aggregation Network (HMA) for single-image super-resolution. Our model combines Fused Convolution with self-attention to better integrate different-level features during deep feature extraction. Additionally, inspired by images' inherent hierarchical structural similarity, we introduce a Grid Attention Block for modeling long-range dependencies. The proposed network enhances multi-level structural similarity modeling by combining sparse attention with window attention. For the super-resolution task, we also designed a pre-training strategy specifically to stimulate the model's potential capabilities further. Extensive experiments demonstrate that our proposed method outperforms state-of-the-art approaches on benchmark datasets for single-image super-resolution tasks. 
{
    \small
    \bibliographystyle{ieeenat_fullname}
    \bibliography{ref}

\begin{thebibliography}{45}
\providecommand{\natexlab}[1]{#1}
\providecommand{\url}[1]{\texttt{#1}}
\expandafter\ifx\csname urlstyle\endcsname\relax
  \providecommand{\doi}[1]{doi: #1}\else
  \providecommand{\doi}{doi: \begingroup \urlstyle{rm}\Url}\fi

\bibitem[Bachmann et~al.(2022)Bachmann, Mizrahi, Atanov, and Zamir]{MultiMAE}
Roman Bachmann, David Mizrahi, Andrei Atanov, and Amir Zamir.
\newblock Multimae: Multi-modal multi-task masked autoencoders.
\newblock In \emph{Computer Vision -- ECCV 2022}, pages 348--367, Cham, 2022.
  Springer Nature Switzerland.

\bibitem[Bevilacqua et~al.(2012)Bevilacqua, Roumy, Guillemot, and
  Alberi-Morel]{set5}
Marco Bevilacqua, Aline Roumy, Christine Guillemot, and Marie~Line
  Alberi-Morel.
\newblock Low-complexity single-image super-resolution based on nonnegative
  neighbor embedding.
\newblock 2012.

\bibitem[Bilecen and Ayazoglu(2023)]{Bilecen_2023_CVPR}
Bahri~Batuhan Bilecen and Mustafa Ayazoglu.
\newblock Bicubic++: Slim, slimmer, slimmest - designing an industry-grade
  super-resolution network.
\newblock In \emph{Proceedings of the IEEE/CVF Conference on Computer Vision
  and Pattern Recognition (CVPR) Workshops}, pages 1623--1632, 2023.

\bibitem[Cao et~al.(2022)Cao, Liang, Zhang, Li, Zhang, Wang, and Gool]{refsr}
Jiezhang Cao, Jingyun Liang, Kai Zhang, Yawei Li, Yulun Zhang, Wenguan Wang,
  and Luc~Van Gool.
\newblock Reference-based image super-resolution with deformable attention
  transformer.
\newblock In \emph{Computer Vision -- ECCV 2022}, pages 325--342, Cham, 2022.
  Springer Nature Switzerland.

\bibitem[Chen et~al.(2021)Chen, Wang, Guo, Xu, Deng, Liu, Ma, Xu, Xu, and
  Gao]{Chen_2021_CVPR}
Hanting Chen, Yunhe Wang, Tianyu Guo, Chang Xu, Yiping Deng, Zhenhua Liu, Siwei
  Ma, Chunjing Xu, Chao Xu, and Wen Gao.
\newblock Pre-trained image processing transformer.
\newblock In \emph{Proceedings of the IEEE/CVF Conference on Computer Vision
  and Pattern Recognition (CVPR)}, pages 12299--12310, 2021.

\bibitem[Chen et~al.(2023)Chen, Wang, Zhou, Qiao, and Dong]{Chen_2023_CVPR}
Xiangyu Chen, Xintao Wang, Jiantao Zhou, Yu Qiao, and Chao Dong.
\newblock Activating more pixels in image super-resolution transformer.
\newblock In \emph{Proceedings of the IEEE/CVF Conference on Computer Vision
  and Pattern Recognition (CVPR)}, pages 22367--22377, 2023.

\bibitem[Chen et~al.(2024)Chen, Wu, Zamfir, Zhang, Zhang, Timofte, Yang,
  et~al.]{chen2024ntire_sr}
Zheng Chen, Zongwei Wu, Eduard-Sebastian Zamfir, Kai Zhang, Yulun Zhang, Radu
  Timofte, Xiaokang Yang, et~al.
\newblock Ntire 2024 challenge on image super-resolution (x4): Methods and
  results.
\newblock In \emph{Computer Vision and Pattern Recognition Workshops}, 2024.

\bibitem[Conde et~al.(2023)Conde, Choi, Burchi, and Timofte]{SWIN2SR}
Marcos~V. Conde, Ui-Jin Choi, Maxime Burchi, and Radu Timofte.
\newblock Swin2sr: Swinv2 transformer for compressed image super-resolution
  and restoration.
\newblock In \emph{Computer Vision -- ECCV 2022 Workshops}, pages 669--687,
  Cham, 2023. Springer Nature Switzerland.

\bibitem[Dai et~al.(2019)Dai, Cai, Zhang, Xia, and Zhang]{san}
Tao Dai, Jianrui Cai, Yongbing Zhang, Shu-Tao Xia, and Lei Zhang.
\newblock Second-order attention network for single image super-resolution.
\newblock In \emph{Proceedings of the IEEE/CVF Conference on Computer Vision
  and Pattern Recognition (CVPR)}, 2019.

\bibitem[Deng et~al.(2009)Deng, Dong, Socher, Li, Li, and Fei-Fei]{5206848}
Jia Deng, Wei Dong, Richard Socher, Li-Jia Li, Kai Li, and Li Fei-Fei.
\newblock Imagenet: A large-scale hierarchical image database.
\newblock In \emph{2009 IEEE Conference on Computer Vision and Pattern
  Recognition}, pages 248--255, 2009.

\bibitem[Du and Tian(2024)]{10225288}
Weizhi Du and Shihao Tian.
\newblock Transformer and gan-based super-resolution reconstruction network for
  medical images.
\newblock \emph{Tsinghua Science and Technology}, 29\penalty0 (1):\penalty0
  197--206, 2024.

\bibitem[Ebrahimi and Vrscay(2007)]{selfsimilarity}
Mehran Ebrahimi and Edward~R. Vrscay.
\newblock Solving the inverse problem of image zooming using ``self-examples''.
\newblock In \emph{Image Analysis and Recognition}, pages 117--130, Berlin,
  Heidelberg, 2007. Springer Berlin Heidelberg.

\bibitem[Gu and Dong(2021)]{gu2021interpreting}
Jinjin Gu and Chao Dong.
\newblock Interpreting super-resolution networks with local attribution maps.
\newblock In \emph{Proceedings of the IEEE/CVF Conference on Computer Vision
  and Pattern Recognition}, pages 9199--9208, 2021.

\bibitem[Huang et~al.(2015)Huang, Singh, and Ahuja]{Huang_2015_CVPR}
Jia-Bin Huang, Abhishek Singh, and Narendra Ahuja.
\newblock Single image super-resolution from transformed self-exemplars.
\newblock In \emph{Proceedings of the IEEE Conference on Computer Vision and
  Pattern Recognition (CVPR)}, 2015.

\bibitem[Kim et~al.(2016)Kim, Lee, and Lee]{Kim_2016_CVPR}
Jiwon Kim, Jung~Kwon Lee, and Kyoung~Mu Lee.
\newblock Accurate image super-resolution using very deep convolutional
  networks.
\newblock In \emph{Proceedings of the IEEE Conference on Computer Vision and
  Pattern Recognition (CVPR)}, 2016.

\bibitem[Kornblith et~al.(2019)Kornblith, Norouzi, Lee, and
  Hinton]{pmlr-v97-kornblith19a}
Simon Kornblith, Mohammad Norouzi, Honglak Lee, and Geoffrey Hinton.
\newblock Similarity of neural network representations revisited.
\newblock In \emph{Proceedings of the 36th International Conference on Machine
  Learning}, pages 3519--3529. PMLR, 2019.

\bibitem[Ledig et~al.(2017)Ledig, Theis, Huszar, Caballero, Cunningham, Acosta,
  Aitken, Tejani, Totz, Wang, and Shi]{Ledig_2017_CVPR}
Christian Ledig, Lucas Theis, Ferenc Huszar, Jose Caballero, Andrew Cunningham,
  Alejandro Acosta, Andrew Aitken, Alykhan Tejani, Johannes Totz, Zehan Wang,
  and Wenzhe Shi.
\newblock Photo-realistic single image super-resolution using a generative
  adversarial network.
\newblock In \emph{Proceedings of the IEEE Conference on Computer Vision and
  Pattern Recognition (CVPR)}, 2017.

\bibitem[Li et~al.(2021)Li, Lu, Qian, Lu, Zhang, and Jia]{li2021efficient}
Wenbo Li, Xin Lu, Shengju Qian, Jiangbo Lu, Xiangyu Zhang, and Jiaya Jia.
\newblock On efficient transformer-based image pre-training for low-level
  vision.
\newblock \emph{arXiv preprint arXiv:2112.10175}, 2021.

\bibitem[Li et~al.(2023)Li, Fan, Xiang, Demandolx, Ranjan, Timofte, and
  Van~Gool]{Li_2023_CVPR}
Yawei Li, Yuchen Fan, Xiaoyu Xiang, Denis Demandolx, Rakesh Ranjan, Radu
  Timofte, and Luc Van~Gool.
\newblock Efficient and explicit modelling of image hierarchies for image
  restoration.
\newblock In \emph{Proceedings of the IEEE/CVF Conference on Computer Vision
  and Pattern Recognition (CVPR)}, pages 18278--18289, 2023.

\bibitem[Liang et~al.(2021)Liang, Cao, Sun, Zhang, Van~Gool, and
  Timofte]{SwinIR}
Jingyun Liang, Jiezhang Cao, Guolei Sun, Kai Zhang, Luc Van~Gool, and Radu
  Timofte.
\newblock Swinir: Image restoration using swin transformer.
\newblock In \emph{Proceedings of the IEEE/CVF International Conference on
  Computer Vision (ICCV) Workshops}, pages 1833--1844, 2021.

\bibitem[Lim et~al.(2017)Lim, Son, Kim, Nah, and
  Mu~Lee]{Lim_2017_CVPR_Workshops}
Bee Lim, Sanghyun Son, Heewon Kim, Seungjun Nah, and Kyoung Mu~Lee.
\newblock Enhanced deep residual networks for single image super-resolution.
\newblock In \emph{Proceedings of the IEEE Conference on Computer Vision and
  Pattern Recognition (CVPR) Workshops}, 2017.

\bibitem[Liu et~al.(2021)Liu, Lin, Cao, Hu, Wei, Zhang, Lin, and
  Guo]{Liu_2021_ICCV}
Ze Liu, Yutong Lin, Yue Cao, Han Hu, Yixuan Wei, Zheng Zhang, Stephen Lin, and
  Baining Guo.
\newblock Swin transformer: Hierarchical vision transformer using shifted
  windows.
\newblock In \emph{Proceedings of the IEEE/CVF International Conference on
  Computer Vision (ICCV)}, pages 10012--10022, 2021.

\bibitem[Lu et~al.(2022{\natexlab{a}})Lu, Li, Liu, Huang, Zhang, and
  Zeng]{Lu_2022_CVPR}
Zhisheng Lu, Juncheng Li, Hong Liu, Chaoyan Huang, Linlin Zhang, and Tieyong
  Zeng.
\newblock Transformer for single image super-resolution.
\newblock In \emph{Proceedings of the IEEE/CVF Conference on Computer Vision
  and Pattern Recognition (CVPR) Workshops}, pages 457--466,
  2022{\natexlab{a}}.

\bibitem[Lu et~al.(2022{\natexlab{b}})Lu, Li, Liu, Huang, Zhang, and
  Zeng]{lu2022transformer}
Zhisheng Lu, Juncheng Li, Hong Liu, Chaoyan Huang, Linlin Zhang, and Tieyong
  Zeng.
\newblock Transformer for single image super-resolution.
\newblock In \emph{Proceedings of the IEEE/CVF conference on computer vision
  and pattern recognition}, pages 457--466, 2022{\natexlab{b}}.

\bibitem[Luo et~al.(2016)Luo, Li, Urtasun, and Zemel]{NIPS2016_c8067ad1}
Wenjie Luo, Yujia Li, Raquel Urtasun, and Richard Zemel.
\newblock Understanding the effective receptive field in deep convolutional
  neural networks.
\newblock In \emph{Advances in Neural Information Processing Systems}. Curran
  Associates, Inc., 2016.

\bibitem[Maaz et~al.(2023)Maaz, Shaker, Cholakkal, Khan, Zamir, Anwer, and
  Shahbaz~Khan]{EdgeNeXt}
Muhammad Maaz, Abdelrahman Shaker, Hisham Cholakkal, Salman Khan, Syed~Waqas
  Zamir, Rao~Muhammad Anwer, and Fahad Shahbaz~Khan.
\newblock Edgenext: Efficiently amalgamated cnn-transformer architecture
  for mobile vision applications.
\newblock In \emph{Computer Vision -- ECCV 2022 Workshops}, pages 3--20, Cham,
  2023. Springer Nature Switzerland.

\bibitem[Martin et~al.(2001)Martin, Fowlkes, Tal, and Malik]{bsd100}
D. Martin, C. Fowlkes, D. Tal, and J. Malik.
\newblock A database of human segmented natural images and its application to
  evaluating segmentation algorithms and measuring ecological statistics.
\newblock In \emph{Proceedings Eighth IEEE International Conference on Computer
  Vision. ICCV 2001}, pages 416--423 vol.2, 2001.

\bibitem[Matsui et~al.(2017)Matsui, Ito, Aramaki, Fujimoto, Ogawa, Yamasaki,
  and Aizawa]{matsui2017sketch}
Yusuke Matsui, Kota Ito, Yuji Aramaki, Azuma Fujimoto, Toru Ogawa, Toshihiko
  Yamasaki, and Kiyoharu Aizawa.
\newblock Sketch-based manga retrieval using manga109 dataset.
\newblock \emph{Multimedia Tools and Applications}, 76:\penalty0 21811--21838,
  2017.

\bibitem[Mei et~al.(2020)Mei, Fan, Zhou, Huang, Huang, and Shi]{Mei_2020_CVPR}
Yiqun Mei, Yuchen Fan, Yuqian Zhou, Lichao Huang, Thomas~S. Huang, and Honghui
  Shi.
\newblock Image super-resolution with cross-scale non-local attention and
  exhaustive self-exemplars mining.
\newblock In \emph{Proceedings of the IEEE/CVF Conference on Computer Vision
  and Pattern Recognition (CVPR)}, 2020.

\bibitem[Mei et~al.(2021)Mei, Fan, and Zhou]{nlsa}
Yiqun Mei, Yuchen Fan, and Yuqian Zhou.
\newblock Image super-resolution with non-local sparse attention.
\newblock In \emph{Proceedings of the IEEE/CVF Conference on Computer Vision
  and Pattern Recognition (CVPR)}, pages 3517--3526, 2021.

\bibitem[Su et~al.(2023)Su, Gan, Chen, Yin, and Chen]{9992208}
Jian-Nan Su, Min Gan, Guang-Yong Chen, Jia-Li Yin, and C.~L.~Philip Chen.
\newblock Global learnable attention for single image super-resolution.
\newblock \emph{IEEE Transactions on Pattern Analysis and Machine
  Intelligence}, 45\penalty0 (7):\penalty0 8453--8465, 2023.

\bibitem[Timofte et~al.(2017)Timofte, Agustsson, Van~Gool, Yang, and
  Zhang]{Timofte_2017_CVPR_Workshops}
Radu Timofte, Eirikur Agustsson, Luc Van~Gool, Ming-Hsuan Yang, and Lei Zhang.
\newblock Ntire 2017 challenge on single image super-resolution: Methods and
  results.
\newblock In \emph{Proceedings of the IEEE Conference on Computer Vision and
  Pattern Recognition (CVPR) Workshops}, 2017.

\bibitem[Tu et~al.(2022)Tu, Mei, Ma, and Piccialli]{9829280}
Jingzhi Tu, Gang Mei, Zhengjing Ma, and Francesco Piccialli.
\newblock Swcgan: Generative adversarial network combining swin transformer and
  cnn for remote sensing image super-resolution.
\newblock \emph{IEEE Journal of Selected Topics in Applied Earth Observations
  and Remote Sensing}, 15:\penalty0 5662--5673, 2022.

\bibitem[Wang et~al.(2023)Wang, Gao, Li, Zhang, and Hu]{pmlr-v202-wang23e}
Shaoru Wang, Jin Gao, Zeming Li, Xiaoqin Zhang, and Weiming Hu.
\newblock A closer look at self-supervised lightweight vision transformers.
\newblock In \emph{Proceedings of the 40th International Conference on Machine
  Learning}, pages 35624--35641. PMLR, 2023.

\bibitem[Wang et~al.(2018)Wang, Yu, Wu, Gu, Liu, Dong, Qiao, and
  Change~Loy]{Wang_2018_ECCV_Workshops}
Xintao Wang, Ke Yu, Shixiang Wu, Jinjin Gu, Yihao Liu, Chao Dong, Yu Qiao, and
  Chen Change~Loy.
\newblock Esrgan: Enhanced super-resolution generative adversarial networks.
\newblock In \emph{Proceedings of the European Conference on Computer Vision
  (ECCV) Workshops}, 2018.

\bibitem[Yang et~al.(2019)Yang, Zhang, Tian, Wang, Xue, and Liao]{srcnn}
Wenming Yang, Xuechen Zhang, Yapeng Tian, Wei Wang, Jing-Hao Xue, and Qingmin
  Liao.
\newblock Deep learning for single image super-resolution: A brief review.
\newblock \emph{IEEE Transactions on Multimedia}, 21\penalty0 (12):\penalty0
  3106--3121, 2019.

\bibitem[Yoo et~al.(2023)Yoo, Kim, Lee, Kim, Lee, and Kim]{Yoo_2023_WACV}
Jinsu Yoo, Taehoon Kim, Sihaeng Lee, Seung~Hwan Kim, Honglak Lee, and Tae~Hyun
  Kim.
\newblock Enriched cnn-transformer feature aggregation networks for
  super-resolution.
\newblock In \emph{Proceedings of the IEEE/CVF Winter Conference on
  Applications of Computer Vision (WACV)}, pages 4956--4965, 2023.

\bibitem[Zamfir et~al.(2023)Zamfir, Conde, and Timofte]{Zamfir_2023_CVPR}
Eduard Zamfir, Marcos~V. Conde, and Radu Timofte.
\newblock Towards real-time 4k image super-resolution.
\newblock In \emph{Proceedings of the IEEE/CVF Conference on Computer Vision
  and Pattern Recognition (CVPR) Workshops}, pages 1522--1532, 2023.

\bibitem[Zeyde et~al.(2012)Zeyde, Elad, and Protter]{set14}
Roman Zeyde, Michael Elad, and Matan Protter.
\newblock On single image scale-up using sparse-representations.
\newblock In \emph{Curves and Surfaces}, pages 711--730, Berlin, Heidelberg,
  2012. Springer Berlin Heidelberg.

\bibitem[Zhang et~al.(2022)Zhang, Zhang, Gu, Zhang, Kong, and
  Yuan]{zhang2022accurate}
Jiale Zhang, Yulun Zhang, Jinjin Gu, Yongbing Zhang, Linghe Kong, and Xin Yuan.
\newblock Accurate image restoration with attention retractable transformer.
\newblock \emph{arXiv preprint arXiv:2210.01427}, 2022.

\bibitem[Zhang et~al.(2018)Zhang, Li, Li, Wang, Zhong, and Fu]{Zhang_2018_ECCV}
Yulun Zhang, Kunpeng Li, Kai Li, Lichen Wang, Bineng Zhong, and Yun Fu.
\newblock Image super-resolution using very deep residual channel attention
  networks.
\newblock In \emph{Proceedings of the European Conference on Computer Vision
  (ECCV)}, 2018.

\bibitem[Zhao et~al.(2022)Zhao, Cao, Huang, and Yang]{9779501}
Mo Zhao, Gang Cao, Xianglin Huang, and Lifang Yang.
\newblock Hybrid transformer-cnn for real image denoising.
\newblock \emph{IEEE Signal Processing Letters}, 29:\penalty0 1252--1256, 2022.

\bibitem[Zhou et~al.(2020)Zhou, Zhang, Zuo, and Loy]{ignn}
Shangchen Zhou, Jiawei Zhang, Wangmeng Zuo, and Chen~Change Loy.
\newblock Cross-scale internal graph neural network for image super-resolution.
\newblock In \emph{Advances in Neural Information Processing Systems}, pages
  3499--3509. Curran Associates, Inc., 2020.

\bibitem[Zhou et~al.(2023)Zhou, Li, Guo, Bai, Cheng, and Hou]{zhou2023srformer}
Yupeng Zhou, Zhen Li, Chun-Le Guo, Song Bai, Ming-Ming Cheng, and Qibin Hou.
\newblock Srformer: Permuted self-attention for single image super-resolution.
\newblock In \emph{Proceedings of the IEEE/CVF International Conference on
  Computer Vision}, pages 12780--12791, 2023.

\bibitem[Zontak and Irani(2011)]{5995401}
Maria Zontak and Michal Irani.
\newblock Internal statistics of a single natural image.
\newblock In \emph{CVPR 2011}, pages 977--984, 2011.

\end{thebibliography}
}
\clearpage
\setcounter{page}{1}
\maketitlesupplementary

\section{Training Details}
\subsection{Study on the pre-training strategy}
\label{sec:pretraining}
\begin{figure}[H]
\centering
\includegraphics[width=1.0\columnwidth]{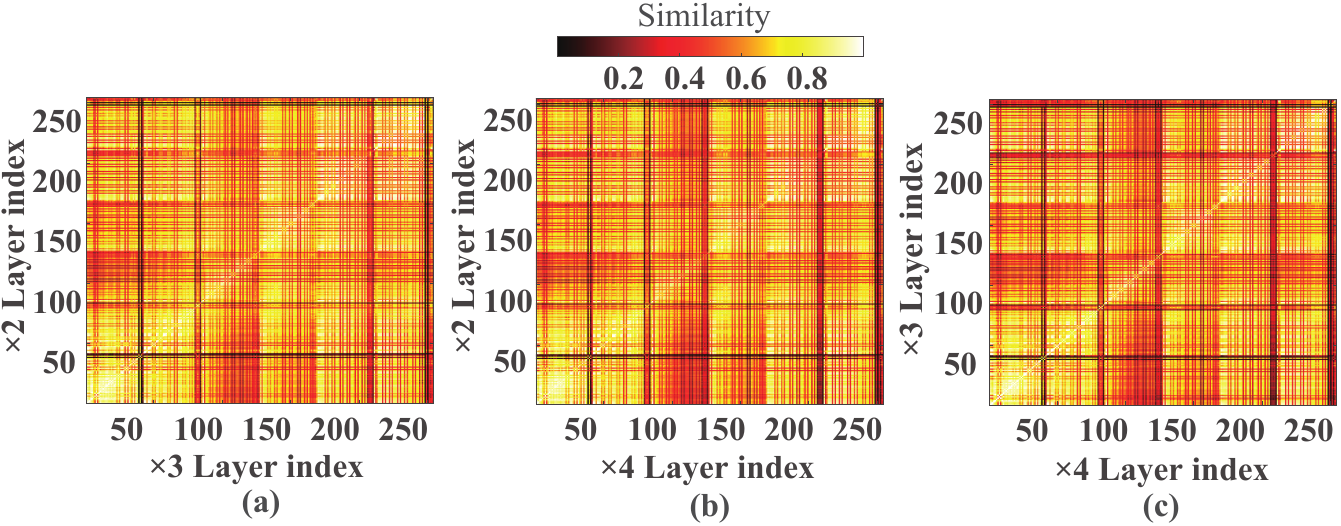}
\caption{(a) CKA similarity map between layers of the $\times$2 SR model and the $\times$3 SR model, (b) CKA similarity map between layers of the $\times$2 SR model and the $\times$4 SR model, (c) CKA similarity map between layers of the $\times$3 SR model and the $\times$4 SR model.}
\label{fig8}
\end{figure}

%
We calculate the interlayer CKA~\cite{pmlr-v97-kornblith19a} similarity in $\times$2 SR, $\times$3 SR, and $\times$4 SR, except for the shallow feature extraction and image reconstruction modules. In \cref{fig8}, we can see that \cref{fig8}(a) and Fig.~\ref{fig8}(c) show high similarity on the diagonal, while \cref{fig8}(b) has a low similarity score on the diagonal. Therefore, we train the $\times$3 SR model after training the $\times$2 SR model as the initial parameter and then use the $\times$3 SR model as the initial parameter of the $\times$2 SR model and the $\times$4 SR model.
We train the model using nine pre-training strategies to test the impact of different pre-training strategies on performance. \cref{tab1} shows the training results, which are evaluated on the Set5~\cite{set5} dataset. We can find that our proposed pre-training strategies can effectively improve the model performance (0.05dB$\sim$0.09dB). It can also be observed that using models with different degradation levels as model initialization parameters has different effects on motivating the model potential. Using the $\times$3 SR model as the initialization parameter for the $\times$2 and the $\times$4 SR models maximizes the model performance. Whereas using the $\times$2 SR model as the initialization parameter of the $\times$4 model, on the contrary, reduces the model performance. This suggests that a suitable pre-training strategy can lead to better performance gains for HMA.

\begin{table}[H]

  \centering
  \scalebox{0.69}{
    \begin{tabular}{|c|c|c|c|c|c|c|c|c|}
    \hline
    \multirow{3}{*}{Scale} & \multicolumn{8}{c|}{Initialization parameters} \\
\cline{2-9}          & \multicolumn{2}{c|}{w/o} & \multicolumn{2}{c|}{$\times$2} & \multicolumn{2}{c|}{$\times$3} & \multicolumn{2}{c|}{$\times$4} \\
\cline{2-9}          & PSNR  & SSIM  & PSNR  & SSIM  & PSNR  & SSIM  & PSNR  & SSIM \\
    \hline
    $\times$2    & 38.84 & 0.9642 & 38.86 & 0.9644 & \textbf{38.95} & \textbf{0.9647} & 38.78 & 0.964 \\
    \hline
    $\times$3    & 35.25 & 0.9342 & \textbf{35.35} & \textbf{0.9346} & 35.27 & 0.9343 & \textbf{35.30} & 0.9345 \\
    \hline
    $\times$4    & 33.26 & 0.9083 & 33.24 & 0.9081 & \textbf{33.38} & \textbf{0.9086} & 33.25 & 0.9083 \\
    \hline
    \end{tabular}%
    }
    \caption{Quantitative results of HMA PSNR (dB) on $\times$4 SR using different pre-training strategies.}\label{tab1}%
\end{table}%

\section{Analysis of Model Complexity}
\label{Complexity}

We experiments to analyze Grid Attention Block (GAB) and Fused Attention Block (FAB). We also compare our method with the Transformer-based method SwinIR. The $\times$4 SR performance on Urban100 is reported and the number of Multiply-Add operations is computed when the input size is 64$\times$64. Note that the pre-training technique is not used for all models in this section.

we use SwinIR with a window size of 16 as a baseline to study the computational complexity of the proposed GAB and FAB. As shown in \cref{tab7}, our GAB obtains performance gains by finitely increasing parameters and Multi-Adds. It proves the effectiveness and efficiency of the proposed modules. In addition, FAB brings better performance at the same time although it brings more parameters and Multi-Adds.
\begin{table}[H]
  \centering
    \begin{tabular}{c|ccc}
    \hline
    Method & \#Params. & \#Multi-Adds. & PSNR\\
    \hline
    SwinIR & 12.1M & 63.8G & 27.81dB\\
    \hline
    w/GAB & 24.4M & 76.9G & 28.37dB\\
    \hline
    w/FCB & 57.6M & 157.0G & 28.30dB\\
    \hline
    Ours  & 69.9M & 170.1G & 28.42dB\\
    \hline
    \end{tabular}%
  \caption{Model complexity comparison of GAB and FAB.}\label{tab7}%
\end{table}%

\section{Visual Comparisons with LAM}
\label{LAM}
\begin{figure*}
\centering
\includegraphics[width=1.0\textwidth]{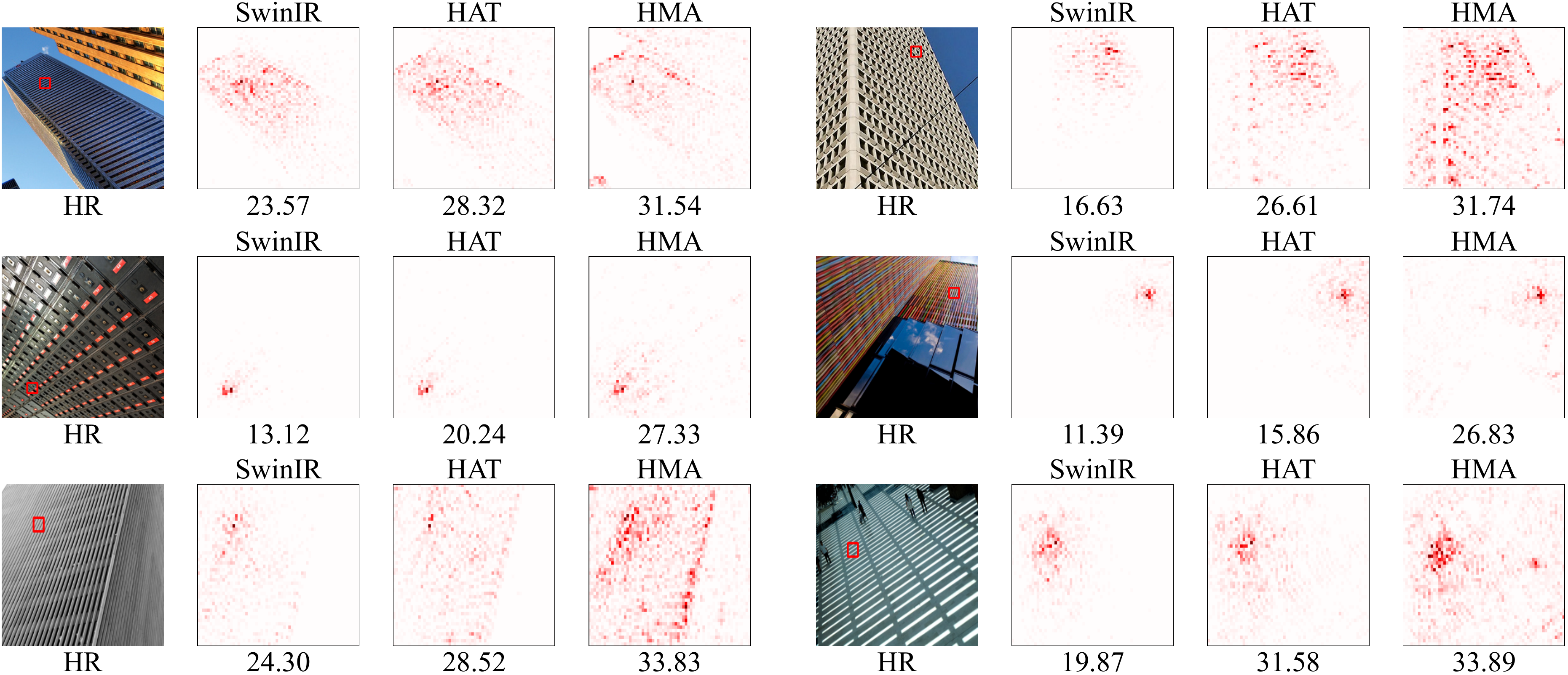}
\caption{Comparison of LAM results between SwinIR, HAT and HMA.}
\label{fig10}
\end{figure*}

We provide visual comparisons with the LAM~\cite{gu2021interpreting} results to compare SwinIR, HAT, and our proposed HMA. The red dots in the LAM results represent the pixels used for reconstructing the patches marked with red boxes in the HR images, and we give the Diffusion Indices (DI) in \cref{fig10} to reflect the range of pixels involved.
In this case, the more pixels are used to recover a specific input block, the wider the distribution of red dots in LAM, and the higher the DI. As shown in \cref{fig10}, both HAT and HMA can effectively extend the effective pixel range compared to the baseline SwinIR, where the pixel range is only clustered in a limited area. Compared to HAT, HMA can extend the range of utilized pixels more widely due to the introduction of the GAB module. Also, for quantitative metrics, HMA obtains much higher DI values than SwinIR and HAT. The visualization results and quantitative evaluation metrics show that HMA can better utilize global information for local area reconstruction. As a result, the method generated by HMA is more capable of generating high-resolution images with better visualization.

\end{document}